\pdfoutput=1

\documentclass[11pt]{article}

\usepackage[preprint]{coling}

\usepackage{times}
\usepackage{latexsym}

\usepackage[T1]{fontenc}
\usepackage[utf8]{inputenc}
\usepackage{amsmath}

\usepackage{microtype}
\usepackage{inconsolata}
\usepackage{graphicx}
\usepackage{subfiles}

\usepackage{booktabs}
\usepackage[inline]{enumitem}
\usepackage{ulem}
\usepackage{accents}

\usepackage[many]{tcolorbox}
\usepackage{pgfplots}
\pgfplotsset{compat=1.17}
\usepackage{colortbl}

\newtcolorbox{textbox}{
    boxrule = 1.5pt,
    colframe = black
}

\title{\textit{Can LLMs Help Create Grammar?}: Automating Grammar Creation for Endangered Languages with In-Context Learning}

\author{Piyapath T Spencer\textsuperscript{1,2} \and Nanthipat Kongborrirak\textsuperscript{1}\\
  \textsuperscript{1}Language and Information Technology Programme, Faculty of Arts, CU, Thailand \\
  \textsuperscript{2}Center for Information and Language Processing (CIS), LMU Munich, Germany \\
  \texttt{linguistics@piyapath.uk}\quad\quad\texttt{prompt.k10@gmail.com}
}

\begin{document}
\maketitle
\begin{abstract}
Yes! In the present-day documenting and preserving endangered languages, the application of Large Language Models (LLMs) presents a promising approach. This paper explores how LLMs, particularly through in-context learning, can assist in generating grammatical information for low-resource languages with limited amount of data. We takes Moklen as a case study to evaluate the efficacy of LLMs in producing coherent grammatical rules and lexical entries using only bilingual dictionaries and parallel sentences of the unknown language without building the model from scratch. Our methodology involves organising the existing linguistic data  and prompting to efficiently enable to generate formal XLE grammar. Our results demonstrate that LLMs can successfully capture key grammatical structures and lexical information, although challenges such as the potential for English grammatical biases remain. This study highlights the potential of LLMs to enhance language documentation efforts, providing a cost-effective solution for generating linguistic data and contributing to the preservation of endangered languages.
\end{abstract}

\section{Introduction}
In linguisitics, the quest for universality \cite{chomsky}, despite its criticism, motivates the comparison of languages to identify common patterns and structures. Many linguists investigate how different languages handle concepts like tense, number, or syntactic structure to identify universal features or constraints.

Within this broader discourse, Lexical-Functional Grammar (LFG) offers a model for understanding the complexities of different information within and across languages. In its different structures, it assumes both variability and universality. For instance, the analysis of a language with a free word order sequence is valid within the LFG framework \cite{simpson}.

\begin{figure}
    \centering
    \includegraphics[width=1\linewidth]{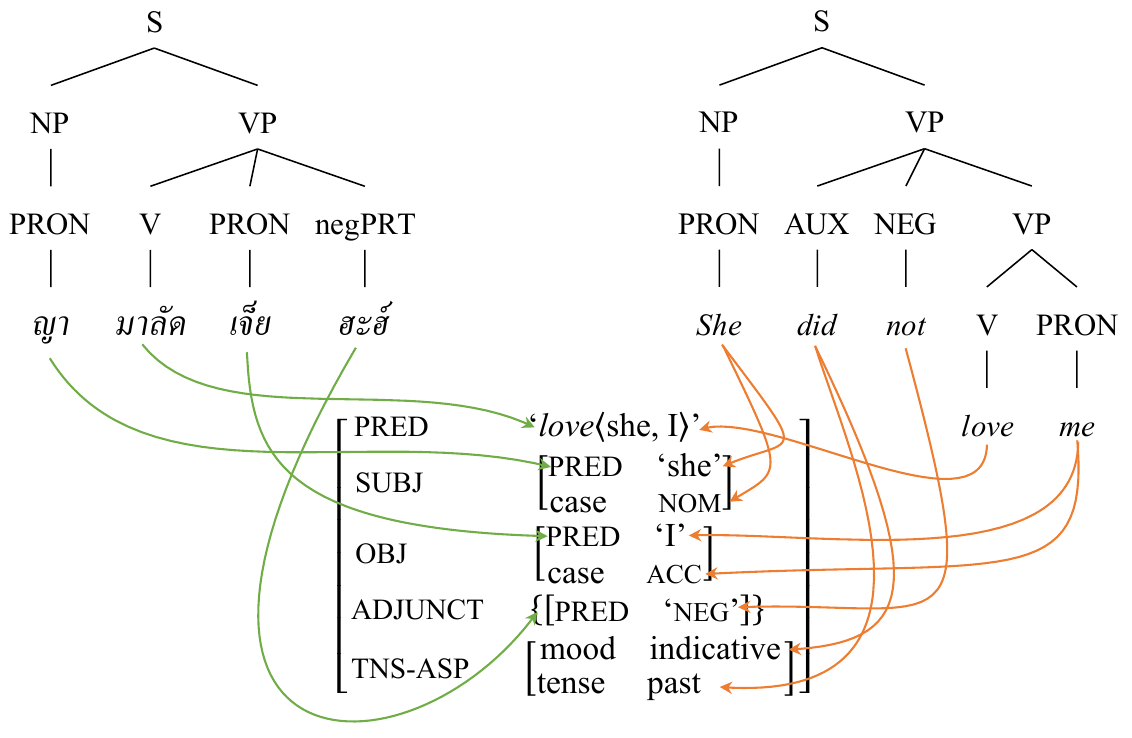}
    \caption{The $\phi$ mapping \cite{Dalrymple2023} from different c(onstituent)-structures of Moklen and English to the same f(unctional)-structure}
    \label{fig:enter-label}
\end{figure}

According to \citet{butt1999}, "a common set of linguistic principles and a commonly agreed upon set of grammatical analyses and features" unify linguistic insights across languages whilst acknowledge the uniqueness of individual languages. 

The XLE (Xerox Linguistic Environment), developed on the basis of LFG \cite{maxwell, documentation}, serves as a powerful tool for the computational implementation of multilingual grammars (grammars across a wide range of languages). However, the 'grammars' are usually manually constructed since it requires the understanding of nuances within a language, which is expensive and time-consuming to produce. This fact makes it difficult to start developing deep grammar for low-resource languages, especially endangered languages. 

The recent development of large language models (LLM) has shown notable capabilities for 'adaptation' \cite{brown2020languagemodelsfewshotlearners, wei2022finetunedlanguagemodelszeroshot} to various tasks and understanding natural language instructions through in-context learning and understanding natural language instructions with in-context learning \cite{brown2020languagemodelsfewshotlearners, openai2024gpt4technicalreport, gemini}. Their linguistic performances and metalinguistic abilities have also been recognised \cite{beguš2023largelinguisticmodelsanalyzing}. This presents an opportunity to use their capabilities in generating grammar and linguistic information for endangered languages, potentially overcoming the need for exhaustive resources and the limitations of manual grammar construction.

This paper proposes a novel approach that exploits LLMs' linguistic competences in English language to generate such coherent formal language as XLE grammar for natural languages that have not been encountered during the model's pre-training. This approach relies solely on bilingual dictionaries and few pairs of  parallel sentences, which reflect the amount of data linguists typically has at hand during the process of documenting endangered languages \cite{spencer-2024-documenting}.

\section{Background and Related Work}
Despite the significant advancements, most existing studies focus on high-resource languages, with limited exploration of how LLMs can be effectively applied to lower-resource or truly low-resource languages. This gap is particularly concerning given the vast number of languages spoken worldwide, many of which lack sufficient digital representation and the risk of extinction.

\noindent{\textbf{On Large Multilingual Grammar Development}}\\
Large multilingual grammar development has seen significant advancements through theoretical grammar models  such as Head-Driven Phrase Structure Grammar (HPSG) and Lexical-Functional Grammar (LFG). Such projects include LinGO \cite{lingo}, Matrix \cite{matrix}, and DELPH-IN \cite{Delph:02:CLE} for HPSG, and the ParGram project \cite{pargram} for LFG. These frameworks often rely on rule-based systems to capture the syntactic and semantic nuances of multiple languages. Their common objective is to create comprehensive grammatical analyses that can be applied across diverse linguistic contexts. However, the manual construction of these grammars demands extensive linguistic expertise and resources, which can be considered as a barrier for low-resource languages.

\noindent{\textbf{On '\textit{Low-Resource}'}}\\
In the fields of artificial intelligence (AI), natural language processing (NLP), and LLMs, the term "low-resource" is surprisingly is a broad. This classification even includes languages like German, Filipino, and some institutional languages that have their own pretrained models, machine translation (MT) systems, and various NLP applications. This disparity arises from the uneven distribution of linguistic resources, with only 14 languages comprising over 90\% of internet content, and English alone accounting for half of all data \cite{content}. In contrast, there are over 7,000 known languages, with approximately half classified as endangered according to \citet{Ethnologue}. Many of these endangered languages are likely to disappear by the end of this century.

When considering endangered languages alongside the NLP concept of low-resource languages, it becomes evident that many of these languages even lacks native speakers in their own society, not to mentioned its minimal or no digital presence. Often, they are primarily spoken languages without a writing system, meaning there may be no written documentation or corpus available. Despite these challenges, linguists have made efforts to document these sorts of languages by, firstly, collecting vocabulary, creating grammar books and dictionaries to preserve their existence of language in the world.

\noindent{\textbf{On Competence and Performance}}\\
Psycholinguistically, a distinction exists between linguistic competence and performance \cite{competence}. Competence refers to the innate ability to control all aspects of a language's structure, ranging from the intricate array of grammatical rules to pragmatic nuances of usage. Performance, on the other hand, pertains to the actual production of language in real-world contexts. This distinction differentiates a critical difference between human linguistic abilities and those of AI, as LLMs possess extensive knowledge of a language due to their rigorous pretraining on large datasets.

When discussing generative AI, the focus is often on how language is created or generated, as suggested by the term itself. Generative AI, including LLMs, thus simulates human linguistic performance \cite{genai}. In the context of linguistics, this raises questions about how language is produced and understood. Over time, benchmarks for LLMs have shown saturation, indicating that models are becoming increasingly powerful and capable of performing complex tasks, as exemplified by models like GPT-4. This leads to comparisons between AI capabilities and human language abilities, prompting inquiries into whether LLMs truly understand language.

Generally, the performance of LLMs is largely determined by both quality and, but much more important, quantity of the data on which they are trained. Whilst LLMs may demonstrate impressive capabilities across various languages, this does not necessarily guarantee a true understanding of those languages. This distinction is particularly evident in the case of English, where LLMs benefit from extensive training on vast and diverse datasets. As a result, their performance in English is often more robust and nuanced compared to their performance in lower-resource languages, which may not have the same level of data availability and linguistic representation. 

\noindent{\textbf{On In-Context Learning}}\\
Recent advancements in LLMs have improved their performance not by training from scratch or requiring extensive datasets, but through prompt engineering that enables in-context learning. This approach allows models to adopt information presented within the context of specific inputs to generate relevant and coherent responses. Various techniques, such as zero-shot, few-shot, and chain-of-thought prompting, have emerged as effective strategies for enhancing LLM capabilities.

In the context of low-resource languages, in-context learning has been employed to refine LLM performance, particularly in machine translation. For instance, \citet{tanzer2024benchmarklearningtranslatenew} utilised dictionaries and grammar books to translate endangered languages with LLMs, establishing benchmarks for evaluation. \citet{zhang-etal-2024-hire}, whose work this paper aims to build upon, extended this approach to cover multiple NLP tasks, evaluating their LINGOLLM methodology across different endangered languages. In contrast to these studies, we seek to advance the application of LLMs in a more radical linguistic task: analysing the grammar of languages under extremely limited conditions where prior data is not applicable.

\section{Language under Study: Moklen Language}
Moklen (ISO 639-3: \texttt{mkm}) is an endangered language estimated to be spoken by fewer than 1,000 people, which constitutes only one-fourth of the total population, most of whom are over 50 years old. The speakers are predominantly scattered across Phang Nga and Phuket in Southern Thailand. Moklen has been influenced by Malay and Thai, the latter being the national language of Thailand. It is classified as one of the Austronesian languages.

Despite attempts by the community to use the Thai alphabet to record and teach Moklen to children, the language remains primarily oral, with no formal written tradition or administrative use. Consequently, there is little to no evidence of the language being used on the internet. Moklen is analysed to have a subject-verb-object (SVO) word order and a nominative-accusative alignment. While this word order is prominent, Moklen also features topicalisation to emphasise certain parts of an utterance. Additionally, it lacks inflectional morphology, meaning that words can be combined in various ways to convey meaning. Grammatical features such as tense, aspect, and number are expressed through content words.

Thorough documentation of the language began in late 2017, leading to the development of a pilot version of a dictionary \cite{dictionary}, following the establishment of a Thai-based orthography system \cite{MkmOrthog}. Nevertheless, there are only a few samples of Moklen sentences, primarily derived from field notes, and limited work has been done on the language. 

Considering all these aspects, Moklen is well-suited for the current study. Firstly, it meets the requirements of the paper's approach, which necessitates bitext and a dictionary with the source language (Moklen) and the target language (English). Secondly, the dictionary is relatively comprehensive, containing words that can be used to express a wide range of concepts. Finally, the isolating nature of Moklen syntax allows for the application of LLMs without the complications of inflectional morphology and complex tokenisation.

\section{Methodology}
\begin{figure*}
    \centering
    \includegraphics[width=0.8\linewidth]{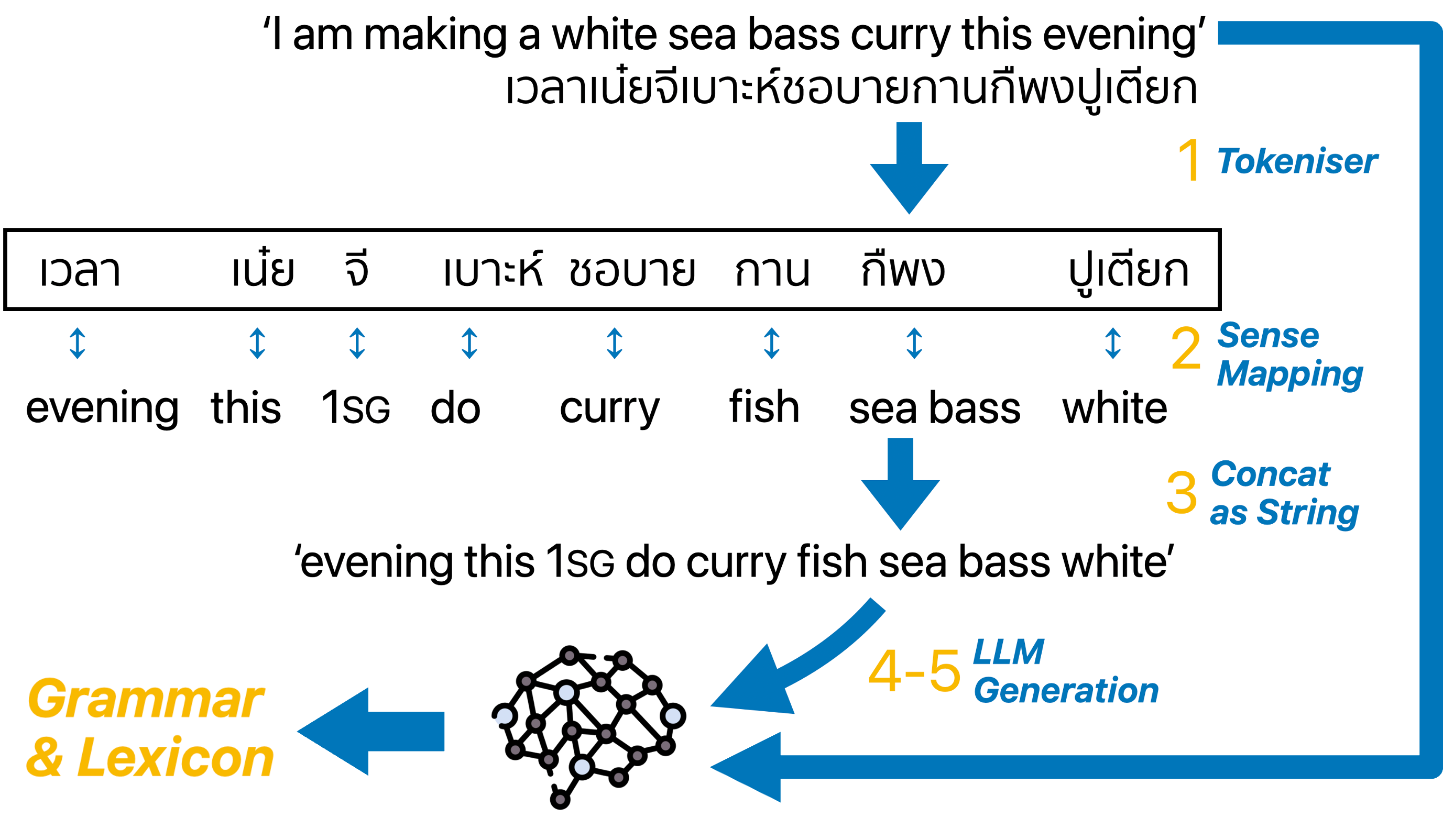}
    \caption{The approach requires each pair of sentences from bitext going through the process of tokenisation, sense mapping, and sentence concatenation prior to include into the prompt to do the task}
    \label{fig2}
\end{figure*}

Our methodology reflects the approach linguists take when analysing grammar and the cognitive aspects of human language. The process comprises five major steps, as shown in \autoref{fig2}:\\
(1) Given the bitext, a dictionary-based tokeniser is used to segment the Moklen text into individual words.\\
(2) For every word in a sentence, we search for the closest match from the dictionary to map the English meaning to the corresponding words in the source language.\\
(3) We concatenate the meanings of each word in each sentence and pair them side-by-side with the original sentences.\\
(4) We prompt a large language model (LLM) with the mapped, tokenised bitext and materials related to creating grammar in XLE format.\\
(5) We then use the generated grammar rules to guide the LLM in generating lexical entries based on the words from the dictionary.\\
The primary data sources for this study include bilingual dictionaries and parallel sentences that pair Moklen with English. The dictionary in this paper developed from \cite{dictionary} serves as a foundational resource, containing approximately 1,000 basic vocabulary items in Moklen along with their English translations. Additionally, a collection of parallel sentences, derived from field notes and existing documentation, will be used to provide contextual examples of grammatical structures in both languages.

\subsection{Tokenisation}
In this step, we create a dictionary-based tokeniser to segment the Moklen text into individual words. This approach is feasible since all words in the Moklen sentences from the bitext exist in the dictionary. However, tokenising Moklen presents unique challenges due to its high frequency of compound words. To address this, a longest-match strategy is employed where the tokeniser first looks for the closest and largest item in the dictionary before considering smaller units. This prevents over-tokenisation and ensures that the true meaning of each sentence is accurately represented.

For example, the Moklen word \textit{maklaw}, meaning 'to speak', is a compound form derived from the root word \textit{klaw}. If tokenised incorrectly, it could be split into two different words: \textit{ma} (meaning 'horse') and \textit{klaw} (meaning 'to speak'). This misinterpretation could lead to a sentence being incorrectly understood as "The horse is speaking." Accordingly, the tokeniser will recognise maklaw as a single unit, thereby preserving the correct meaning and enhancing the overall accuracy of the analysis.

\subsection{Sense Mapping}
Once the text is tokenised, we perform sense mapping by searching for the corresponding meaning of each token in the Moklen sentences. This step is crucial for ensuring that the LLM does not attempt to match each token independently, which often results in inaccuracies. Instead, we aim to provide the LLM with contextually relevant mappings.

Each word in Moklen may have multiple meanings, and it is essential for the LLM to select the appropriate meaning based on the context provided. This process assumes that the LLM can consider the various meanings of words and choose the one that best fits the context of the sentence. For instance, in the sentence \textit{tichum boh pong}, which translates to 'The bird is building the nest', the word \textit{pong} has two different senses: 'nest' and 'father'. The LLM should correctly interpret the sentence as 'The bird is building the nest' rather than 'The bird is building/making the father', demonstrating its ability to disambiguate based on context.

\subsection{Generation of Grammar}
After mapping the meanings, we concatenate the meanings of each word and prepare the data for input into the LLM. The underlying idea is that even though the LLM may have no prior knowledge of Moklen, it can employ its logical reasoning capabilities to analyse the provided data. By comparing the shared structures of Moklen and English, the LLM can induce grammatical rules from the context and information given.

For the model generation, there are two approaches to creating the grammar of Moklen. The first approach involves describing the language using natural language, detailing information such as word order, grammatical features, and functions. The second approach is to generate the grammar in a formal language, specifically XLE, by creating rules and lexical entries using a specific template and syntax\footnote{Henceforth, \textit{grammar} refers to formal rules and lexical entries in XLE.}. Whilst we favour the second approach as the end product of this paper, the natural language description can be a good indicator to understand how the LLM interprets and analyses the language data. 

\section{Experiment}
\subsection{Experimental Setup}    
We ran all experiments on OpenAI's API-based model \texttt{gpt-4o-mini-2024-07-18} and keep a rather low temperature at 0.1 across tasks. The benchmark data, code, and model generations can be found in the supplementary material.

Despite being a single task, we also perform the task under different experimental settings, varying the kind of retrieved context that are provided in the prompt. See \autoref{sec:appendix-prompt} for full details. The types of context include:

    \noindent\textbf{No context (-):} Apart from Moklen sentences, the model is told only that Moklen is a language spoken around Southern Thailand and given no reference material. This measures the base model's zero-shot capabilities and validates that they have effectively learned zero Moklen during pretraining.

    \noindent\textbf{Bitext Context (B):} For each sentence, we provide an English translation into the prompt.

    \noindent\textbf{Tokenised Context (T):} We experiment with two different ways. First, for each sentence, we provide tokens ($T^0$). Second, for each word in each Moklen sentence, we map them with the English definition no matter how many senses they are into the prompt ($T^D$).

    \noindent\textbf{Concatenated Sentence Context (C):} For each sentence, we take the translation of each word to concatenate as one string, though they may be nonsensical or ungrammatical (since we need the incorrect ones on purpose!). It is assumed that this type of string will reflect how word order in Moklen works as opposed to English.

    \noindent\textbf{Example Context (E):} The model is provided with an example of how the English language is implemented with XLE.

    \noindent\textbf{Self-Explanation Context (S):} Prior to the generation, the model is given the relevant data to analyse and describe in natural language, and take it as a guideline when generating the grammar.

For all contexts, the model is retrieval-augmented by providing with the XLE documentation that describes how to create a grammar and a Moklen dictionary except for no context condition, storing in a separate datastore \cite{asai2024reliableadaptableattributablelanguage}.

\subsection{Evaluation Measures}
Evaluating the quality of generated grammars is inherently an complex task, as there is no single correct approach to grammar construction. Consequently, the evaluation must consider both quantitative and qualitative aspects of the model’s performance and the output.

    \noindent\textbf{Translation.} We follow \citet{tanzer2024benchmarklearningtranslatenew} and \citet{zhang-etal-2024-hire} in that translation performance is used as an indirect indicator of the model’s understanding of linguistic structures, albeit not explicitly. Thus, this task is the starting point to selecting the best combination of contexts to proceed to the more sophisticated step. We will evaluate the model's performance on Moklen-to-English translation tasks, using an additional set of 40 parallel sentences. The quality of translations will be assessed using several metrics including a traditional BLEU \cite{papineni-etal-2002-bleu}, ROUGE \cite{ganesan2018rouge20updatedimproved}, METEOR \cite{banerjee-lavie-2005-meteor}, chrF \cite{popovic-2015-chrf}, and BERTScore \cite{zhang2020bertscoreevaluatingtextgeneration}. This very task will serve as an experimental ablation to determine the most effective context settings for further steps. Apart from this, English-to-Moklen translation will be conducted both before and after generating the grammar to determine the impact of the generated grammar on translation performance.

    \noindent\textbf{Grammar Rule Accuracy.} Never before had Moklen been developed its grammar using XLE. Hence, there is no XLE grammar for Moklen; we attempt to create one based on \citet{mkm-vp} and served as a gold standard. The model will be prompted to generate the generated grammar rules which will be qualitatively evaluated. Ideally, the rules must demonstrate the capacity to accurately model all possible sentence structures in Moklen, including the full range of parts-of-speech categories. Specifically, the grammar should cover syntactic constructions, word order, and morphological agreement, ensuring that all core grammatical elements of Moklen are represented.
    
    \noindent\textbf{Lexical Entry and Schemata Accuracy.} The model will generate lexical entries for 100 words from the Moklen dictionary that do not appear in the bitext, primarily based on the grammar rules produced by the model and the information provided in the dictionary. For each lexical entry, there are differences between how different element differently exhibits their functional schemata. This accuracy is also assumed to be resulted by the accuracy of the grammar rules, as well as the understanding of both grammar of the language and XLE architecture. We will manually assess the accuracy of these lexical entries by comparing them to existing dictionary definitions and evaluating their coherence across the generation. we will consider the completeness of the entries, examining whether the model captures various meanings and usages of each word. Ultimately, these lexical entries will be integrated into the XLE system.

\section{Results}
\label{sec:result}

\subsection{Translation} 
\noindent\textbf{Ablation.} We explored 48 possible context combinations (see \autoref{sec
}) and identified 10 particularly noteworthy ones based on representativeness, not necessarily the best performance. These combinations are: \textbf{-}, \textbf{B}, \textbf{T$^0$}, \textbf{B}+\textbf{T$^0$}+\textbf{S}, \textbf{B}+\textbf{T$^D$}, \textbf{B}+\textbf{T$^D$}+\textbf{C}, \textbf{T$^D$}+\textbf{C}+\textbf{S}, \textbf{B}+\textbf{T$^D$}+\textbf{C}+\textbf{E}, \textbf{T$^D$}+\textbf{C}+\textbf{E}+\textbf{S}, and \textbf{B}+\textbf{T$^D$}+\textbf{C}+\textbf{E}+\textbf{S}. Further translation tasks were conducted on these combinations.

Surprisingly, the combination with all contexts did not yield the best performance. Instead, the \textbf{T$^D$} context alone achieved the highest score, with a BERTScore F1 of 0.4904, only slightly better than the second-best. This suggests that \textbf{T$^D$} is more effective when combined with other contexts, outperforming \textbf{T$^0$} in all scenarios. The impact of contexts \textbf{E} and \textbf{S} is relatively equal when combined with other contexts.

\begin{figure}[htbp]
    \centering
    \includegraphics[width=1\linewidth]{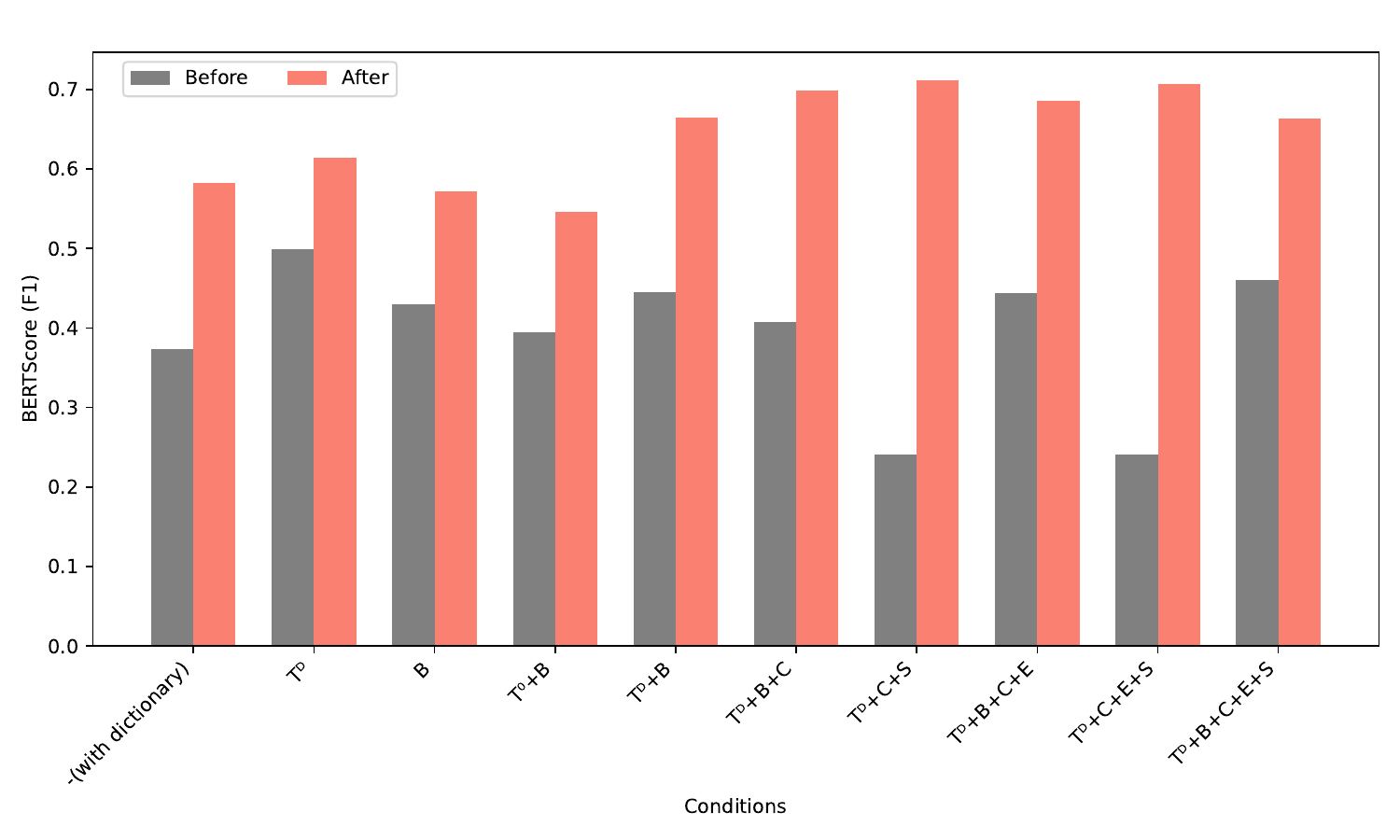}
    \caption{When instructing with grammar, the model performs better across various combinations of context}
    \label{fig:beforeaftergraph}
\end{figure}

\noindent\textbf{English-to-Moklen with Grammar.} 
When incorporating the XLE gold standard grammar into the prompt across all models, we observed an overall improvement in translation performance. This suggests that the inclusion of grammar plays a crucial role in enhancing model output quality.

This finding is consistent with our previous observations. Specifically, the context combination \textbf{T$^D$}+\textbf{C}+\textbf{S} emerged as the most effective, achieving the highest score in nearly all benchmarks with an estimated BERTScore F1 of 0.7110. This result highlights that certain combinations of contexts can significantly enhance translation accuracy.

\noindent\textbf{English-to-Moklen with Grammar.} 
Interestingly, translating from Moklen to English demonstrated better performance compared to the reverse translation (English-to-Moklen). This suggests that the model’s ability to generate English sentences from Moklen inputs is more refined or effective than generating Moklen sentences from English. The implication behind this will be discussed later in the next section.

\begin{figure}[htbp]
    \centering
    \includegraphics[width=1\linewidth]{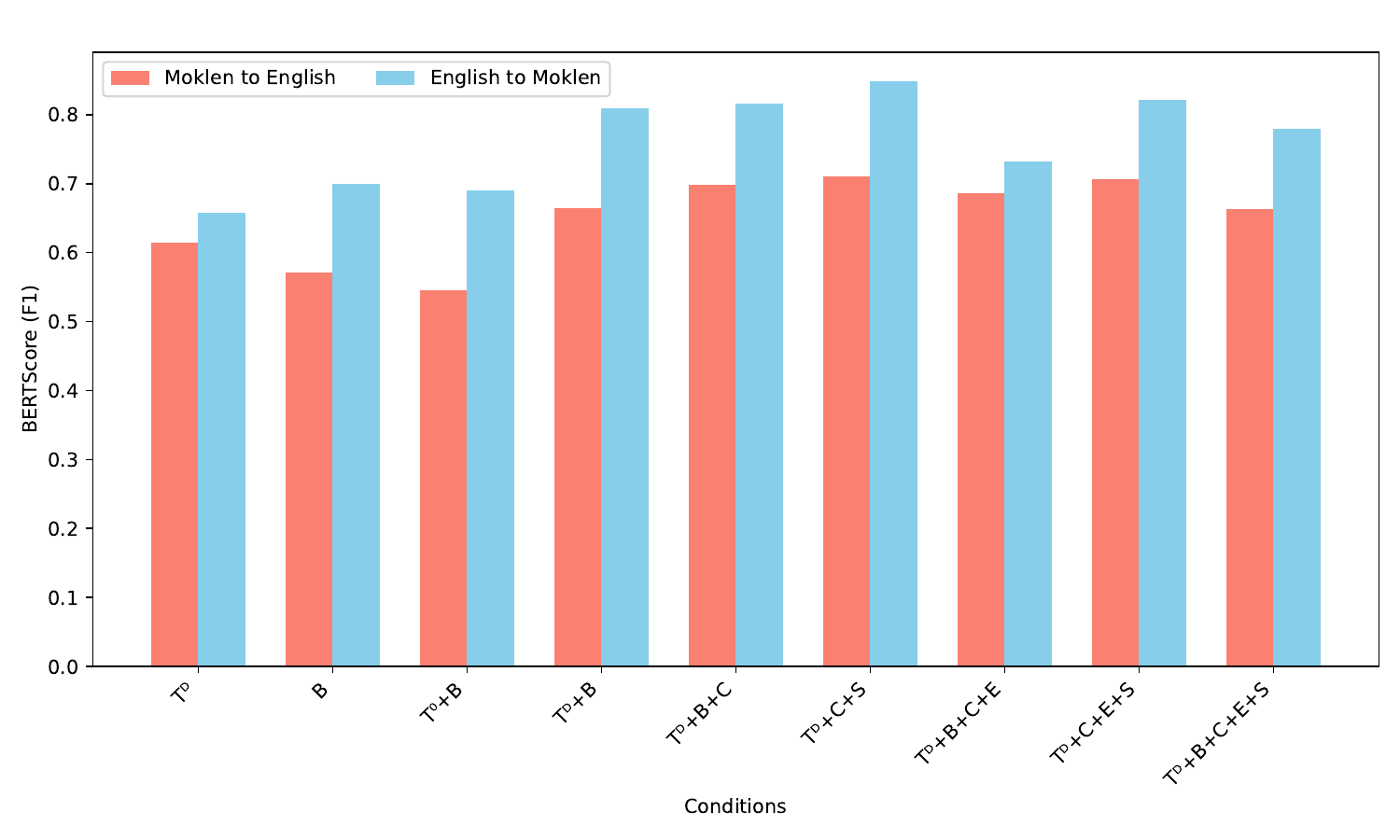}
    \caption{The improvement in scores across contexts leading to the suspicion of the reliability of the metric}
    \label{fig:mkm-eng}
\end{figure}

\subsection{Grammar Rules: Accuracy and Completeness} 
To assess the completeness of the grammar rules, we evaluated whether they could accurately capture all parts of speech in Moklen without overly relying on English examples.

Using the context \textbf{T$^D$}+\textbf{C}+\textbf{S}, the model effectively identified common parts of speech such as nouns, verbs, adjectives, and adverbs. However, some unique English parts of speech, particularly determiners, were incorrectly applied to Moklen grammar. Despite this, the XLE implementation accurately captures most grammatical structures in the language. This discrepancy raises questions that will be explored in the following discussion.

\subsection{Lexical Entries: Accuracy and Coherence} 
We generated lexical entries for 100 Moklen words not present in the bitext and assessed their accuracy by comparing them to existing dictionary definitions. Out of these, 86 entries were deemed accurate and coherent. However, some incomplete lexical entries captured only the more prominent senses of words. For example, the word \textit{data} was defined as a preposition meaning `on' and a noun meaning `the top', but the latter sense was not captured by the model.

\section{Analysis \& Discussion}
There are several critical aspects emerging from the experimental results worth for further discussion.\\
\noindent\textbf{Dictionary Integration.} 
The inclusion of a dictionary in the prompt significantly impacts model performance, even without further contextual modifications. This simple integration yields results that are comparable to or better than combining the dictionary with other contexts. Specifically, the dictionary enhances the model's lexical understanding by grounding it in a structured reference, which is essential when working with languages with minimal resources. For Moklen, a language that lacks morphological inflections and instead uses content words to express grammatical relations (e.g., tense or aspect), dictionary-based lexical grounding proves effective. The performance improvements can be seen in the BERTscores, where the model achieved translation quality approaching that of more morphologically complex languages, despite Moklen's isolating structure. This supports the idea that even minimal data, when correctly utilised, can provide significant gains in language processing tasks.

\begin{figure}[htbp]
    \centering
    \includegraphics[width=1\linewidth]{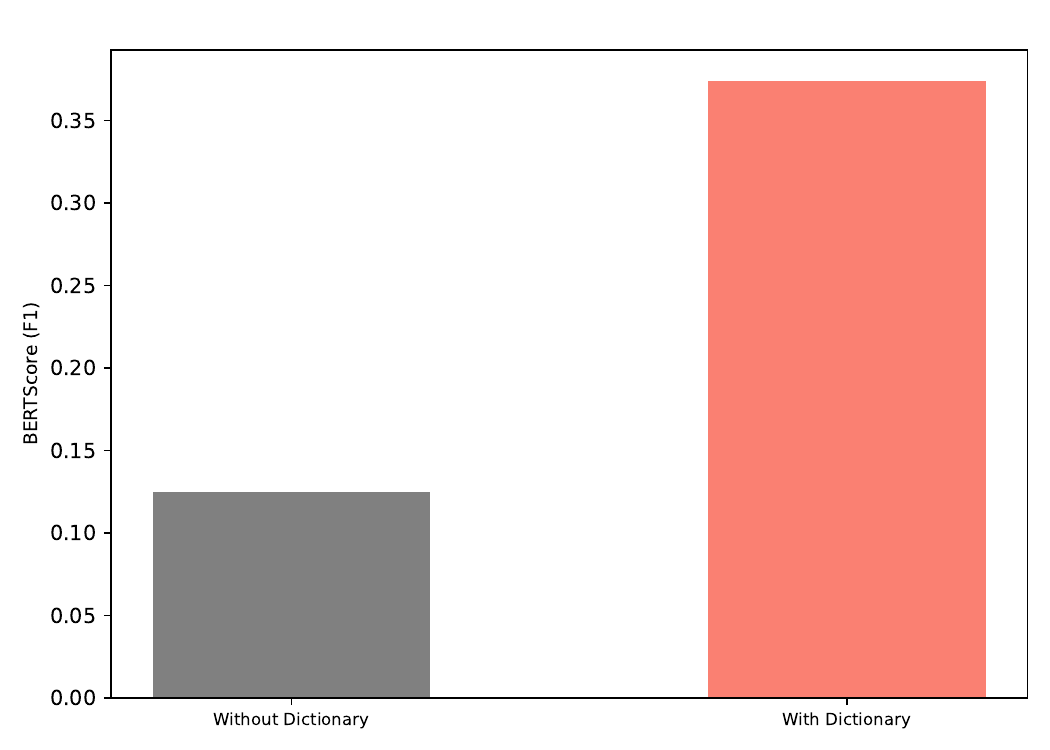}
    \caption{The clear improvement after intergrating dictionary into the prompt in Moklen-to-English translation tasks}
    \label{fig:dictionary}
\end{figure}

\noindent\textbf{Evaluation of Translation Metrics.} 
A key challenge with existing translation metrics like BLEU, ROUGE-L, METEOR, and chrF is their heavy reliance on word forms and syntactic similarity, which aligns poorly with Moklen's structural characteristics. For instance, these metrics work effectively in English, where grammatical markers such as tense are explicitly coded into verb forms. However, Moklen's lack of grammatical tense markers presents a problem, as a single word can express multiple temporal aspects. This reduces the validity of word-form-based metrics for this language, leading to discrepancies between numerical scores and actual translation quality. As an alternative, BERTScore, which prioritizes semantic similarity over word forms, emerges as a better choice for evaluating translations of languages like Moklen. However, even BERTScore falls short in fully capturing the subtleties of Moklen's grammar. To address this, future work should explore the development of a new evaluation metric tailored to languages with similar grammatical features, focusing on meaning representation rather than form.

\noindent\textbf{Model Size Considerations.} 
One critical design decision in this study was the intentional use of a small model. The choice to work with a smaller model was made to demonstrate the potential for generating usable grammar and lexical information with limited computational resources. The success of the small model implies that larger models could yield even better performance, especially in capturing more subtle linguistic nuances or handling rare linguistic phenomena. Given the scalability of LLMs, this opens a pathway for applying similar methods to larger models for endangered languages with more complex syntactic structures or larger datasets.

\noindent\textbf{Hallucinations and Model Inaccuracies.}
In low-resource settings, model hallucinations, particularly in zero-shot translation tasks, were evident. Hallucinations primarily arose when the model was forced to translate or generate sentences with little or no exposure to similar data during training. This was especially pronounced with Moklen, where the use of the Thai script caused confusion. Due to overlap in the lexicons of Thai and Moklen, the model occasionally produced hybridized translations, mistaking Thai words for Moklen ones. Although these inaccuracies were rare, they underscore the importance of refining training data and prompt construction to mitigate such issues. Moving forward, the inclusion of more diverse Moklen data or deliberate disambiguation in the training process may reduce hallucinations and improve model robustness.

However, while these deviations can be problematic, they can also offer unexpected insights for linguists. In some cases, the model's introduction of 'non-Moklen' parts of speech or grammar elements not traditionally associated with the language might suggest an overlooked pattern or nuance. These hallucinations, though initially appearing as errors, can prompt linguists to reconsider their assumptions and explore whether the model has captured subtle relationships or features that were not immediately apparent from a human perspective. For instance, the model's introduction of categories that do not exist formally in Moklen may highlight potential areas of linguistic overlap or underlying structures that deserve further investigation. This kind of speculative output, while not always accurate, provides a thought-provoking dimension to the analysis, encouraging a holistic re-examination of linguistic data.

\noindent\textbf{Potential for XLE Parsing and Beyond.} 
this approach should not be overlooked. XLE's highly sophisticated nature for parsing natural language makes it an ideal framework for documenting endangered languages. However, given the success of this methodology, it is reasonable to assume that the generated grammar could be adapted to other formal frameworks, such as HPSG or dependency grammars, making it widely applicable across different linguistic projects. This flexibility suggests that LLMs, when properly guided, can not only assist in XLE parsing but also contribute to the broader field of computational linguistics by providing insights into the underlying structure of under-documented languages.

\section{Conclusion}
It is evident that an LLM, if sufficient information is provided, can assist linguists in generating linguistic data for such complex and, moreover, tedious tasks. This methodology with careful prompting techniques offers a cost- and time-effective means of obtaining grammatical information for the endangered Moklen language by means of minimal linguistic resources and analogy from English language. By using as less as only bilingual dictionaries and in-context learning, we successfully extracted coherent grammatical rules and lexical entries, thereby contributing to the documentation and preservation of Moklen or, at least, isolating languages in general.

\section*{Limitation}
While this study demonstrates promising results, it is important to acknowledge the limitations of this study. We experimented with only a single language, Moklen, which is an isolating language without grammatical inflections. This typological characteristic may have made it easier for the LLM to decipher the language from a negligible amount of information compared to other languages in the world. We plan to extend this methodology to inclusively experiment with various languages from different typologies and morphologies to assess its broader applicability. Additionally, the potential for English grammatical biases in the model's outputs suggests that further research is needed to develop techniques for minimising cross-linguistic interference in low-resource language modelling.

\bibliography{reference}

\end{document}